\documentclass{article}
\usepackage{spconf,amsmath,graphicx}
\usepackage{url}
\usepackage{multirow} 
\usepackage{booktabs}
\usepackage{subcaption}
\usepackage{hyperref}
\usepackage{colortbl}

\title{BD OPEN LULC MAP: HIGH-RESOLUTION LAND USE LAND COVER MAPPING \& BENCHMARKING FOR URBAN DEVELOPMENT IN DHAKA, BANGLADESH}

\name{
\begin{tabular}{@{}c@{}}
Mir Sazzat Hossain \textsuperscript{\textdagger} \qquad
Ovi Paul \textsuperscript{\textdagger} \qquad
Md Akil Raihan Iftee \qquad
Rakibul Hasan Rajib \\
Abu Bakar Siddik Nayem \qquad
Anis Sarker \qquad
Arshad Momen \quad
Md. Ashraful Amin \\
Amin Ahsan Ali \qquad
AKM Mahbubur Rahman\sthanks{Corresponding author. Email: \href{mailto:akmmrahman@iub.edu.bd}{akmmrahman@iub.edu.bd}}
\end{tabular}
}
\address{Center for Computational \& Data Sciences, Independent University, Bangladesh}

\begin{document}

\maketitle
\def\thefootnote{\textsuperscript{\textdagger}}\footnotetext{These authors contributed equally to this work.}\def\thefootnote{\arabic{footnote}}
\begin{abstract}
Land Use Land Cover (LULC) mapping using deep learning significantly enhances the reliability of LULC classification, aiding in understanding geography, socioeconomic conditions, poverty levels, and urban sprawl. However, the scarcity of annotated satellite data, especially in South/East Asian developing countries, poses a major challenge due to limited funding, diverse infrastructures, and dense populations. In this work, we introduce the \textbf{BD Open LULC Map (BOLM)}, providing pixel-wise LULC annotations across eleven classes (e.g., Farmland, Water, Forest, Urban Structure, Rural Built-Up) for Dhaka metropolitan city and its surroundings using high-resolution Bing satellite imagery (2.22 m/pixel). BOLM spans 4,392 km\textsuperscript{2} (891 million pixels), with ground truth validated through a three-stage process involving GIS experts. We benchmark LULC segmentation using DeepLab V3+ across five major classes and compare performance on Bing and Sentinel-2A imagery. BOLM aims to support reliable deep models and domain adaptation tasks, addressing critical LULC dataset gaps in South/East Asia.

\end{abstract}

\section{Introduction}

Land use/land cover (LULC) changes have gained significant attention due to rapid global ecosystem transformations \cite{abebe2022analysing}. Developing countries in South and East Asia are experiencing substantial LULC changes. Grasslands, woodlands, bushlands, and other vegetation covers are being extensively converted into agricultural and settlement areas to accommodate growing populations. Land cover refers to the physical materials covering the earth's surface, such as forests, mountains, deserts, and water, while land use describes how humans utilize land for socio-economic activities like farming and urban development. Monitoring LULC changes over time helps understand geographical and socio-economic conditions, poverty levels, and urban sprawl. Recently, deep learning methods have effectively extracted LULC information from satellite images \cite{demissie2017land}. In this paper, we introduce the BD Open LULC Map (BOLM) to address the lack of LULC datasets for developing countries.

To effectively apply deep learning methods for LULC in developing countries with similar topographies (Bangladesh, Thailand, Indonesia, Malaysia, India, and Sri Lanka), three major challenges must be addressed. First, there is a scarcity of LULC-annotated data. Large volumes of satellite images with pixel-level LULC annotations are essential for training deep models. Annotated datasets from developed countries \cite{pot, miniFrance, GID} often produce erroneous results when models are applied to developing countries due to differences in building materials, land characteristics, and cultural contexts. Furthermore, UAV data collection, like in OpenEarthMap \cite{xia2023openearthmap}, is costly and limited in developing countries. Thus, high-resolution (HR) satellite LULC annotations are critically needed, especially for Bangladesh.

\begin{figure}[!t]
    \centering
    \includegraphics[width=\linewidth]{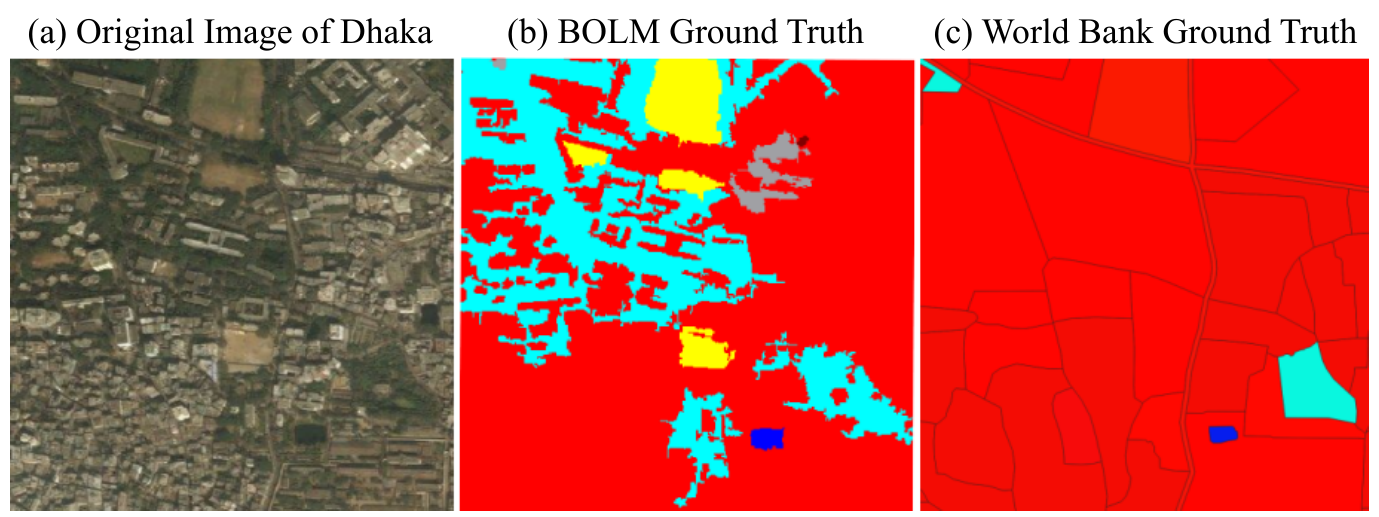}
    \caption{Comparison between World Bank and BOLM (ours) Ground Truth (Here red color represents various kinds of built-up, cyan represents forest/tree cover, yellow represents bare land/meadow and blue represents water)}
    \label{fig:1}
    \vspace{-0.7cm}
\end{figure}

\begin{table*}[!t]
    \centering
    \caption{Summary of the LULC satellite datasets from developing countries}
    \label{tab:1}
    \resizebox{\textwidth}{!}{
    \begin{tabular}{|l|l|l|l|l|l|} 
        \hline
        \rowcolor[gray]{.85}
        Dataset Name & Region & Source & \begin{tabular}[c]{@{}l@{}}Acquisition \\ Time  \end{tabular}& \begin{tabular}[c]{@{}l@{}}GSD \\(Per Pixel)\end{tabular} & \begin{tabular}[c]{@{}l@{}}Area\\(Sq KM)\end{tabular} \\ 
        \hline
        IndiaSat\cite{indianData2}, 2021 & India & \begin{tabular}[c]{@{}l@{}}Landsat-7, Landsat-8 \\and Sentinel-2\end{tabular} & 2016-2019  & 10, 15 m & 162 \\ 
        \hline
        \begin{tabular}[c]{@{}l@{}} OpenEarthMap\cite{xia2023openearthmap},\\2023 \end{tabular} & \begin{tabular}[c]{@{}l@{}}Africa, Asia, Australia, \\Europe, USA \\South America, \\South Asia (Dhaka, Cox's Bazar, Nepal Only)\end{tabular}  & UAV  & 2020 & \begin{tabular}[c]{@{}l@{}} 0.25m –0.5m  \end{tabular} & 799 \\ 
        \hline
        
        \begin{tabular}[c]{@{}l@{}} DynamicEarthNet\cite{toker2022dynamicearthnet},\\2022 \end{tabular} & \begin{tabular}[c]{@{}l@{}}Africa, Australia, Europe, USA \\South America,  China, \\ South Asia (India), East Asia (Thailand)\end{tabular}  & Sentinel 1 \& 2  & 2020-2021 & \begin{tabular}[c]{@{}l@{}} 3m - 10m  \end{tabular} & 707 \\ 
        \hline
        \begin{tabular}[c]{@{}l@{}} LandCoverNet\cite{landcovernet},\\2020 \end{tabular} & \begin{tabular}[c]{@{}l@{}}Africa, Asia, Australia, \\Europe,North America \\and South America\end{tabular}  & Sentinel-2 & 2018 & 10 m & 12,905 \\ 
        \hline
        Mauro et al.\cite{viet}, 2020 & Hanoi, Vietnam  & Landsat-5,7,8 & 1989-2019 & 30 m & 3,050 \\
        \hline

        \begin{tabular}[c]{@{}l@{}} World bank \cite{waterdata},\\2021 \end{tabular} & Dhaka  & \begin{tabular}[c]{@{}l@{}}QuickBird, Pleiades,\\Landsat~and Sentinel-2\end{tabular} & 2006,2017  & 5m-30m & 1,034 \\ 
        \hline
        
        \textbf{BOLM(Ours)} & \textbf{Dhaka} & \textbf{Bing} & \textbf{2019} & \textbf{2.22 m} & \textbf{4,392} \\
        \hline
    \end{tabular}}
    \vspace{-1.5em}
\end{table*}

Second, as shown in Table~\ref{tab:1}, most datasets from this region feature low-resolution images (10m, 15m, 30m), which are inadequate for dense, unstructured urban areas. Low-resolution data and patch-based annotations induce errors in deep learning models. For example, India's datasets \cite{indianData2} lack sufficient resolution and detail for accurate pixel-level LULC mapping. The World Bank's urban mapping of Dhaka \cite{waterdata} lacks necessary details for precise annotations.

Third, many deep learning methods rely solely on RGB channels, achieving good performance compared to traditional approaches  \cite{reis2008analyzing}. However, HR RGB data is not frequently updated in developing countries \cite{bingMap}. Conversely, satellite data (e.g., Sentinel-2/3, Landsat) are regularly updated and offer multi-spectral channels, which can be leveraged using indices like NDVI, NDWI, and RVI \cite{ndvi}. Despite their lower resolution, these channels can enhance LULC mapping when combined with DL models. The key contributions of this paper are:

\begin{itemize}
    \vspace{-0.3em}
    \item Introduction of the BD Open LULC Map (BOLM), a high-resolution, pixel-by-pixel annotated dataset covering 4,392 sq km of Dhaka's urban and rural areas, using Bing satellite imagery (2.22 m/pixel).
    \vspace{-0.3em}
    \item Benchmarking of state-of-the-art deep learning algorithms (DeepLabV3+ \cite{chen2018encoderdecoder}) on both low-resolution (Sentinel-2A) and high-resolution (Bing) data.
    \vspace{-0.3em}
    \item Comprehensive analysis of model performance using different Sentinel-2A channels and indices (NDVI, NDWI, RVI), and comparison with HR data results.
    \vspace{-0.3em}
    \item Conclusive recommendations on the applicability of BOLM dataset for developing countries in South and East Asia, for both Sentinel and Bing/Google imagery.
\end{itemize}

\section{Existing Datasets and Limitations}
\label{sec:existing}
Several high-quality LULC datasets exist, mainly from developed countries. The Potsdam and Vaihingen datasets \cite{pot} cover German regions with five classes: building, low-veg, tree, car, background. The Gaofen (GID) dataset \cite{GID} spans China with classes like forest, built-up, farmland, meadow, and water. MiniFrance \cite{miniFrance} provides urban and rural scene annotations for China and France. FloodNet \cite{floodnetData} includes ten LULC classes related to floods in Texas and Louisiana.

Turning to developing countries, India’s  IndiaSat \cite{indianData2} uses Sentinel and Landsat imagery for land cover classification. Africa’s Mpologoma Catchment dataset \cite{bunyangha2021past} (Uganda) and Libokemkem district dataset \cite{demissie2017land} (Ethiopia) cover land changes over time. LandCoverNet \cite{landcovernet} offers global LULC data, while a Hanoi study \cite{viet} tracks urbanization from 1989–2019.  Despite these, high-quality, pixel-wise annotated LULC data remains scarce in developing regions, and existing datasets (Table \ref{tab:1}) have several limitations:

\begin{itemize}
    \item Most datasets use low-resolution imagery (10m–30m), making it hard to distinguish small urban features like houses, roads, and vegetation. Unlike structured cities, developing regions lack clear zoning, complicating land cover classification.  
    \vspace{-0.3em}
    \item Deep learning models trained on one region often fail in others due to geographic and cultural variations \cite{developingWorldsensors}. Differences in building materials, landscapes, and urban structures cause domain shifts, requiring high-resolution $(\sim2m)$ datasets with precise pixel-wise annotations.
    \vspace{-0.3em}
    \item LULC datasets from Bangladesh have coarse annotations, leading to misclassified pixels and degraded model performance. For instance, the World Bank’s urban mapping of Dhaka merges key categories (e.g., construction sites, airports, and industrial areas), covers only urban regions, and is not publicly available. As Figure \ref{fig:1} illustrates, our annotations offer a finer level of detail compared to the World Bank's ground truth.
\end{itemize}

\section{Dataset Creation}  
\label{sec:dataset} 
The study focuses on the Dhaka division, centered around Dhaka city, covering a 117 km × 124 km region from N24°23'04", E89°57'26" (upper left corner) to N23°22'13", E91°00'26" (lower right). 

\subsection{Sources of Satellite Data}  
This study utilizes two image sources: i) Bing imagery (2.22 m/pixel) for ground truth annotation and high-resolution segmentation, and ii) Sentinel-2A data for experiments using various band combinations and index images as inputs.

\begin{figure}[htbp]
    \centering
    \includegraphics[width=0.45\textwidth]{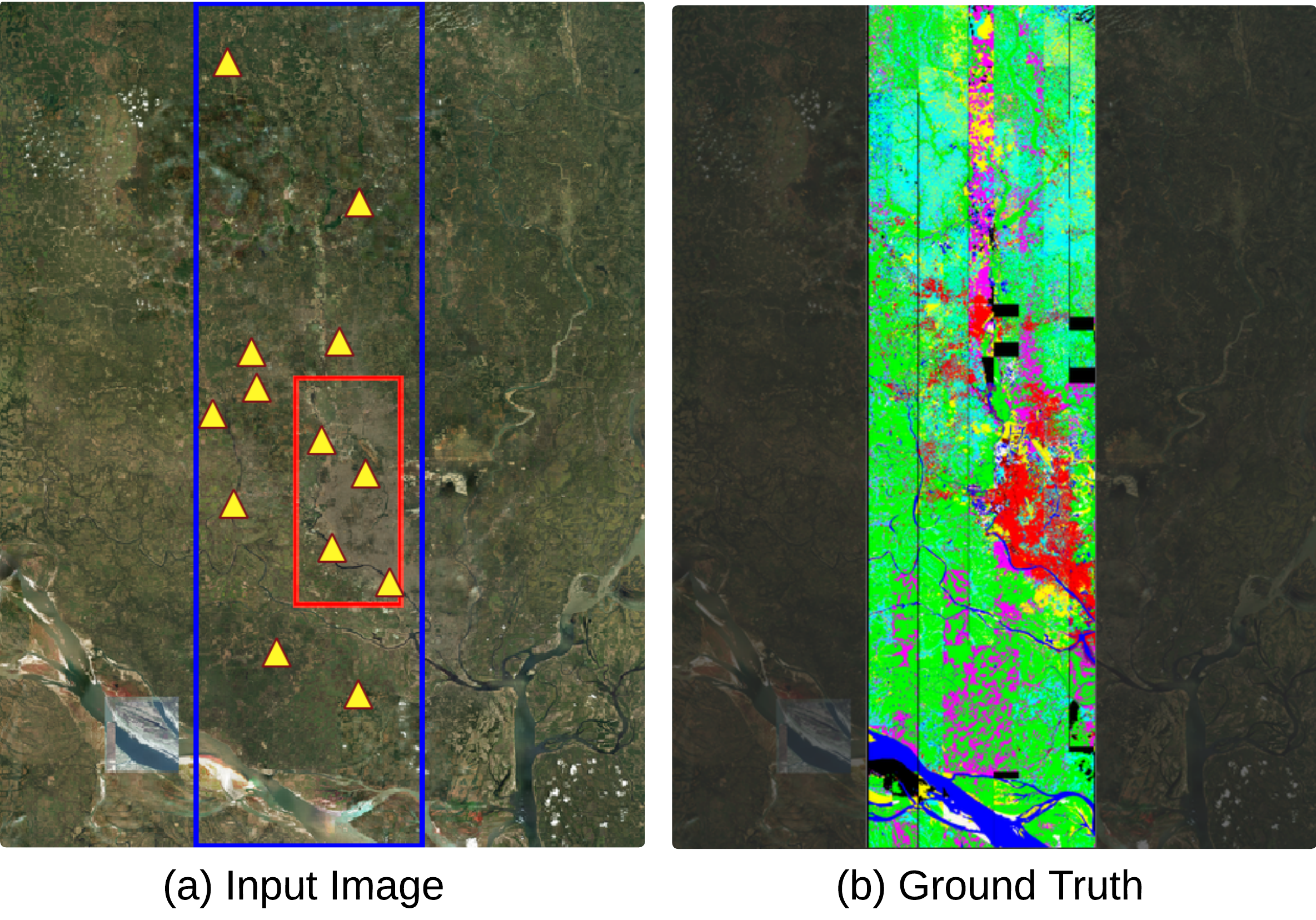}
    \caption{(a) is the input image from the bing used and (b) is the corresponding ground truth.}
    \label{bing}
    \vspace{-0.6em}
\end{figure}

\textbf{Bing RGB: }Bing provides high-resolution aerial imagery (2.22 m/pixel) at 17 zoom levels \cite{bing}. Acquired April 20, 2019, the dataset has a $48906\times47256$ resolution in TIFF (RGB) format.  Figure \ref{bing} shows the full image, blue rectangle marking the annotated area (Dhaka and surroundings) and the red rectangle highlighting the test region within Dhaka city.

\textbf{Sentinel-2A: }Sentinel-2A data, acquired on April 11, 2019, at 04:38:29 UTC (Tile ID: "L1C\_T45QZG\_A016175
\_20190411T043829"), includes 12 spectral bands with resolutions from 10m to 60m per pixel. The red (band 2), green (band 3), blue (band 4), and NIR (band 8) have 10m resolution, while Coastal Aerosol (band 1), Water Vapor (band 9), and SWIR Cirrus (band 10) have 60m. NIR and SWIR bands are at 20m resolution.  

SWIR combinations help analyze moisture and distinguish minerals. False Color Infrared (FCI) (NIR, red, green) aids vegetation detection, while Atmospheric Penetration (ATM) (two SWIR, one NIR) enhances classification clarity \cite{doi:10.1080/10106040408542323}. NDVI assesses vegetation health, RVI estimates vegetation water content, and NDWI detects water stress in plants, with wetter vegetation yielding higher values.
\vspace{-0.5em}

\subsection{Ground Truth Annotation Process}

The Bing image (2.22 m/pixel) covers 4,392 sq km in Dhaka, totaling 891 million pixels, and it was annotated into eleven LULC classes (see Table \ref{Stage-3pixeldisagree} in the supplementary material).

The large Bing image (48,906 × 47,256 pixels) was divided into a 33x33 grid, resulting in 1,089 sub-images (1,500 × 1,500 pixels). Sub-images from each grid column were placed into a distinct folder, with each folder containing 33 images. Our study focused on Dhaka city and its surrounding area, leading us to annotate folders 8 through 19 (12 folders total), which correspond to the region within the blue rectangle in Figure \ref{bing}. A total of 24 annotators (two per folder) used eCognition software and a rule-based approach to classify polygons. Multi-resolution segmentation was applied, merging adjacent polygons of the same class before exporting the final ground truth. Figures \ref{fig:2} and \ref{fig:4} illustrate the process (see supplementary material).

To ensure accuracy, annotators received training on eCognition usage, rule-set creation, and segmentation. Experts clarified class distinctions, and a tutorial video was provided for reference (footnote \ref{video}).

\footnotetext[1]{Annotation tutorial video: \url{https://www.youtube.com/watch?v=psQwvRDxuTo}\label{video}}

\vspace{-0.7em}

\subsection{Annotation Validation and Data Reliability}
To ensure annotation reliability, we followed a three-stage validation process. In Stage 1, two annotator groups independently labeled the dataset, with a pixel-wise agreement matrix highlighting in Table \ref{Stage-3pixeldisagree} (check the supplementary material) disagreements in all classes except Brick and Road. The highest agreement was observed for the Forest class (95.24\%), while the Road class had the lowest (88.18\%). Stage 2 focused on refining disagreement regions (200 × 200 pixels) through manual corrections, significantly reducing inconsistencies. Finally, Stage 3 involved GIS experts validating the dataset, achieving over 99\% agreement for most classes. The expert-labeled pixels were adopted as the final ground truth, ensuring high-quality annotations.

\vspace{-0.7em}

\subsection{In Situ Visual Assessment}  
To enhance annotation reliability, an on-site validation was conducted in the Dhaka Division. The team visited 13 diverse locations, covering urban, rural, agricultural, industrial, and natural landscapes. Using a high-resolution camera, Gaia GPS, and field notes, they compared dataset annotations with real-world observations.  

\begin{figure}
\centering
  \centering
  \includegraphics[width=\linewidth]{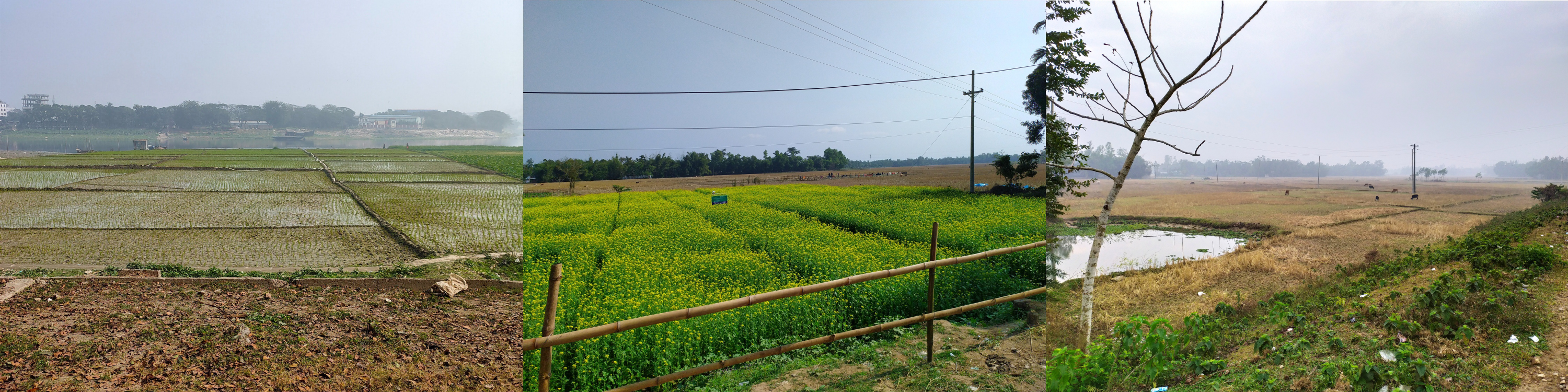}
  \caption{Various Regions on the Dhaka Division}
  \label{dhaka_farm}
  \vspace{-1.2em}
\end{figure}

Land cover characteristics—including infrastructure, vegetation, water bodies, and farmland—were documented with photographs and GPS coordinates for precise alignment. Figure 
 \ref{bing} shows the visited locations, while Figure \ref{dhaka_farm} presents sample images illustrating LULC variations.  

 \vspace{-0.7em}

\subsection{Conversion from 11 Classes to 6 Classes}  
\begin{figure}[!t]
\centering
\begin{subfigure}{0.48\textwidth}
  \centering
  \includegraphics[width=1\linewidth]{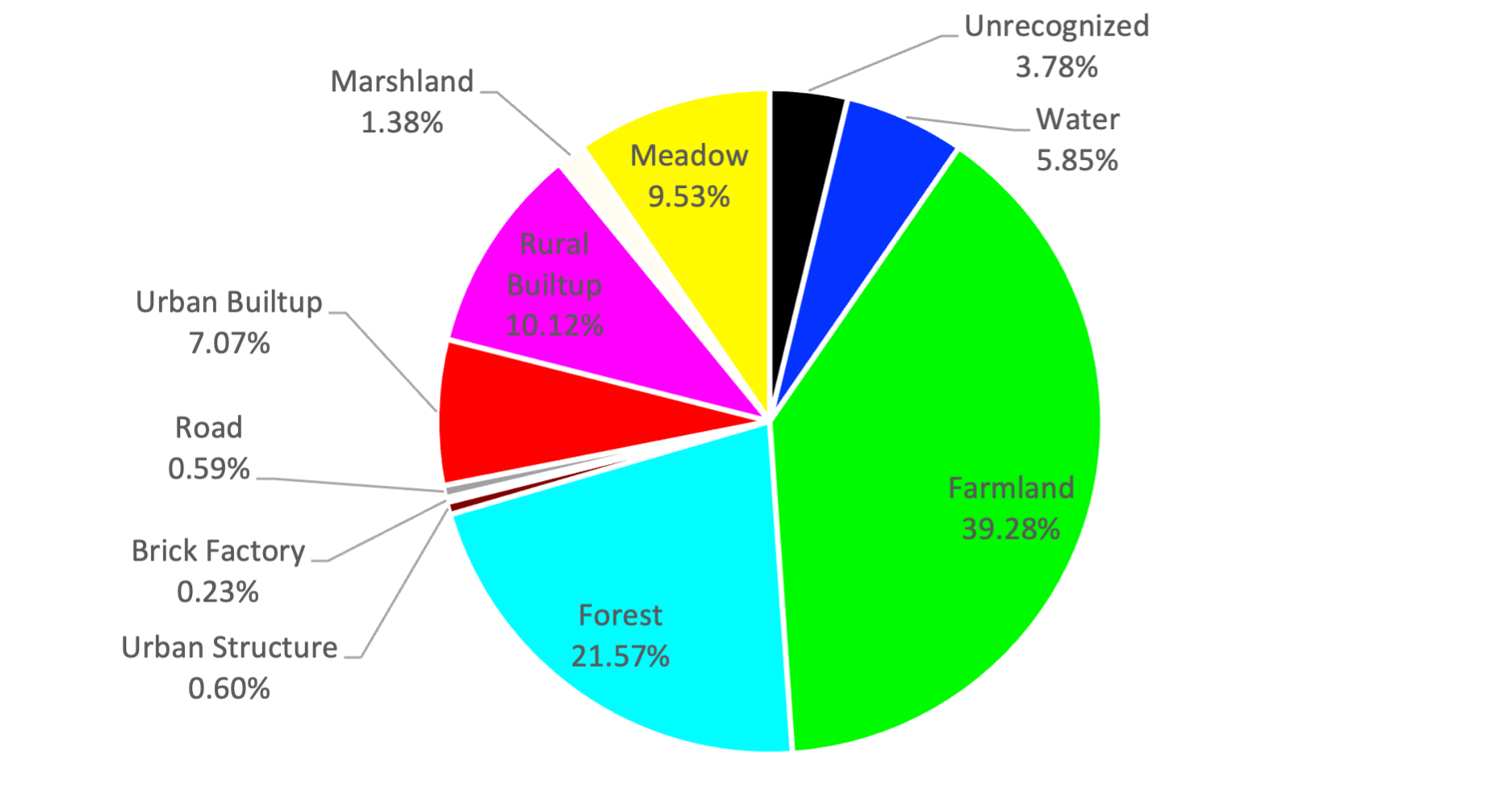}
  \caption{Area Percentage of 11 Class}
  \label{fig:11classpercent}
\end{subfigure}\\
\begin{subfigure}{.48\textwidth}
  \centering
  \includegraphics[width=1\linewidth]{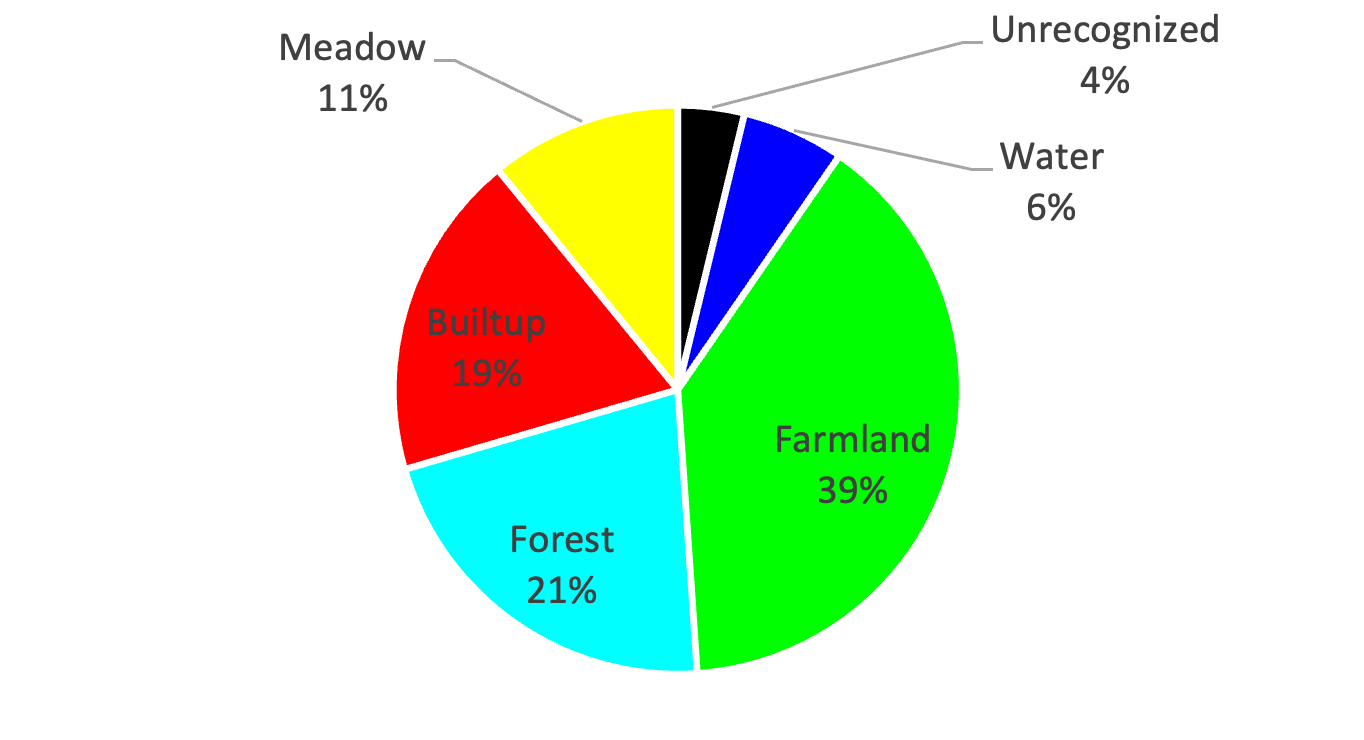}
  \caption{Area Percentage of 6 Class}
  \label{fig:6classpercent}
\end{subfigure}
\caption{Area Percentage of the 11 class and 6 class dataset.}
\label{fig:classpercent}
\vspace{-1.5em}
\end{figure}

Due to class imbalance (Figure \ref{fig:11classpercent}), we merged eleven classes into six broader categories (Table \ref{Stage-3pixeldisagree}) for better balance in experiments. Farmland, Water, and Forest remained unchanged, while Urban Structure, Rural Built-up, Urban Built-up, and Brick Factory merged into Built-up. Meadow and Marshland were combined into Meadow. The final class distribution is: Farmland (39\%), Water (6\%), Forest (21\%), Built-up (19\%), Meadow (11\%), and Unrecognized (3.78\%) (see Figure \ref{fig:6classpercent}).

\vspace{-0.5em}

\section{Image processing for satellite data}
\label{sec:image_processing}

\begin{figure}[htbp]
    \centering
    \includegraphics[width=\linewidth]{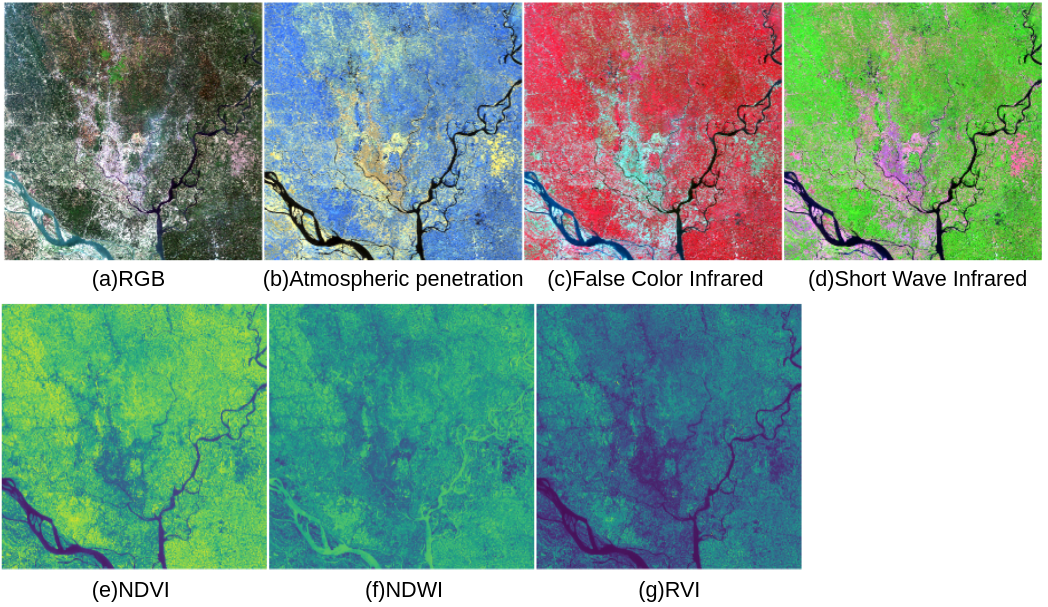}
    \caption{Combination and Index images generated from Sentinel-2 multispectral bands}
    \label{fig:6}
    \vspace{-0.3em}
\end{figure}

\textbf{Sentinel-2A Index and Combination Images: } Sentinel-2A provides 12 spectral bands, which we combine for segmentation tasks. Since different channels capture diverse land textures, we utilize multiple combinations from Sentinel-2A data. Index images are crucial for LULC classification. The RGB representations of these index and combination images are shown in Figure \ref{fig:6}. Given the varying resolutions of Sentinel-2A channels, we upsampled them to 10 m/pixel using bilinear interpolation. The 3-channel index and combination images, listed in Table \ref{tab:sentinel_2_index}, were generated using QGIS.

\begin{table}
\centering
\caption{Sentinel-2 Index and Combination RGB Channel Formation}
\label{tab:sentinel_2_index}

\begin{tabular}{llll}
\hline
Image Type & Channel-1 & Channel-2 & Channel-3 \\ \hline
\multicolumn{1}{l|}{RGB}        & B4(Red)   & B3(Green) & B2(Red)   \\
\multicolumn{1}{l|}{ATM}        & B12(SWIR) & B11(SWIR) & B8(NIR)   \\
\multicolumn{1}{l|}{FCI}        & B8(NIR)   & B4(Red)   & B3(G)      \\
\multicolumn{1}{l|}{SWI}        & B12(SWIR) & B8(NIR)   & B4(R)      \\
\multicolumn{1}{l|}{NDVI}       & NDVI      & NDVI      & NDVI       \\
\multicolumn{1}{l|}{NDWI}       & NDWI      & NDWI      & NDWI      \\
\multicolumn{1}{l|}{RVI}        & RVI       & RVI       & RVI    \\ \hline
\end{tabular}
\end{table}

\noindent \textbf{Bing RGB Image: }The Bing image, captured at a 17-zoom level (2.22 m/pixel), serves as a high-quality ground truth for accurate LULC segmentation. Its high resolution enables precise annotation and improves model performance. However, the large data size increases processing and training time. Despite this challenge, its clarity benefits both manual annotation and deep learning-based segmentation.
\section{Experimental section}
\label{sec:experiments}

\textbf{Implementation details: }For benchmarking, we utilized DeepLabV3+ \cite{chen2018encoderdecoder}, an advanced version of DeepLabV3 with an encoder-decoder structure and atrous convolution for improved semantic segmentation. The input and ground truth images are split into 513×513 patches for DeepLab V3+ training. Some ground truth regions are difficult to label and remain unrecognized, with unrecognized pixels assigned black in both input and ground truth. Training uses a sliding window with 75\% overlap in both directions, while testing uses non-overlapping patches. The Dhaka city region (red rectangle) served as the test set, while surrounding areas (blue rectangle) were used for training shown in  figure \ref{bing}. We computed IoU and F1 scores from confusion matrices based on model predictions and ground truth for test image pixels. Accuracy was not used due to potential bias from imbalanced LULC class samples. The experimental settings are as follows:

 \textbf{LULC with Index Images:} Three-channel input images are created using NDVI, NDWI, and RVI. Training images are generated with a 513×513 sliding window, overlapping 75\% in both directions. The batch size is 8, and training runs for 25 epochs. Testing uses non-overlapping patches.
 
 \textbf{LULC with Channel-Combined Images:} The same sliding window approach is used for training, with a batch size of 8 and 25 epochs, using DeepLabV3+. Testing is performed without overlapping.
    
\textbf{LULC with Full Bing RGB:} The full Bing image is used, following the same training procedure as above, with DeepLabV3+ and no overlap during testing.
\section{Results}
\label{sec:results}
\subsection{Index images}
Among three index images, NDWI achieves the highest IoU (0.20) and F1 score (0.32) for forests and performs best for water (IoU: 0.24, F1: 0.51). NDVI outperforms for built-up and farmland, while RVI has a slightly better F1 score for meadow. On average, NDVI provides the best segmentation for five classes (Avg IoU: 0.30, Avg F1: 0.45). Built-up, farmland are segmented with high precision using index images.


\begin{table}[h]
    \centering
    \caption{Scores for 5 classes of all the image types.}
    \label{tab:sentinel_2_dhaka}
    \resizebox{\linewidth}{!}{%
    \begin{tabular}{lccccccc}
        \toprule
        \textbf{} & \textbf{Metric} & \textbf{Forest} & \textbf{Built-Up} & \textbf{Water} & \textbf{Farmland} & \textbf{Meadow} & \textbf{Avg} \\ 
        \midrule
        
        \multirow{2}{*}{NDVI} 
         & IoU & 0.17 & \textbf{0.52} & \textbf{0.68} & \textbf{0.35} & \textbf{0.51} & \textbf{0.30} \\
         & F1  & 0.17 & \textbf{0.68} & 0.27 & \textbf{0.51} & 0.50 & \textbf{0.45} \\ 
        \midrule
        
        \multirow{2}{*}{NDWI} 
         & IoU & \textbf{0.20} & 0.38 & 0.24 & 0.34 & 0.28 & 0.28 \\
         & F1  & \textbf{0.32} & 0.55 & \textbf{0.38} & 0.51 & 0.50 & 0.43 \\ 
        \midrule
        
        \multirow{2}{*}{RVI} 
         & IoU & 0.13 & 0.50 & 0.17 & 0.31 & 0.36 & 0.29 \\
         & F1  & 0.22 & 0.66 & 0.29 & 0.47 & \textbf{0.52} & 0.43 \\ 
        \midrule
        \midrule
        
        \multirow{2}{*}{ATM} 
         & IoU & 0.17 & \textbf{0.56} & 0.29 & \textbf{0.48} & 0.19 & 0.33 \\
         & F1  & 0.29 & \textbf{0.72} & 0.45 & \textbf{0.64} & 0.31 & 0.48 \\ 
        \midrule
        
        \multirow{2}{*}{FCI} 
         & IoU & 0.20 & 0.53 & 0.30 & 0.46 & 0.17 & 0.33 \\
         & F1  & 0.33 & 0.69 & 0.45 & 0.62 & 0.29 & 0.47 \\ 
        \midrule
        
        \multirow{2}{*}{SWI} 
         & IoU & \textbf{0.22} & 0.56 & \textbf{0.32} & 0.46 & 0.22 & \textbf{0.35} \\
         & F1  & \textbf{0.35} & 0.71 & \textbf{0.48} & 0.62 & 0.37 & \textbf{0.50} \\ 
        \midrule
        
        \multirow{2}{*}{SentinelRGB} 
         & IoU & 0.18 & 0.54 & 0.30 & 0.46 & \textbf{0.25} & 0.34 \\
         & F1  & 0.30 & 0.70 & 0.46 & 0.63 & \textbf{0.40} & 0.49 \\
        \midrule
        \midrule
        
        \multirow{2}{*}{BingRGB} 
         & IoU & 0.33 & 0.58 & 0.48 & 0.57 & 0.26 & 0.44 \\
         & F1  & 0.49 & 0.73 & 0.65 & 0.72 & 0.41 & 0.60 \\
    \bottomrule
    \end{tabular}}
\end{table}

For the farmland class, NDVI, RVI both achieved an F1 score of 0.79, but NDVI had a higher IoU (0.35). As NDWI detects water/moisture content, its superior water detection (IoU: 0.24) was expected. NDVI performed best for built-up areas (IoU: 0.52, F1: 0.68), while RVI was best for meadow (IoU: 0.52, F1: 0.68). NDVI and RVI had the same average F1 score, but NDVI had the highest average IoU (0.30).

\vspace{-0.5em}


\subsection{Channel Combination Images}

We analyze the results of combined images for Sentinel-2A from Table~\ref{tab:sentinel_2_dhaka}. SWI outperforms ATM in Forest segmentation with an IoU of 0.22. For Built-up, both ATM and SWI achieve an IoU of 0.56, but SWI attains higher accuracy. Water regions are better detected with SWI, while ATM performs best for Farmland IoU. Meadow segmentation achieves the highest IoU (0.22) with SWI. On average, SWI attains the highest IoU (0.35) and F1 score (0.50) across all classes.

SentinelRGB exhibits slightly lower performance than SWI. Built-up and Farmland are accurately segmented across all compositions, with ATM achieving the best IoU (0.56) and F1 (0.72) for Built-up and IoU (0.48), F1 (0.64) for Farmland. Forest and Water segmentation is superior with SWI, while Meadow achieves IoU (0.25) and F1 (0.40) with SentinelRGB. Overall, SWI achieves the best average IoU and F1 score among four compositions and three index images derived from Sentinel-2A data.

SWI scores the highest in Forest (0.22 IoU). Built-up IoU is identical for ATM and SWI. SWI leads in Water segmentation, ATM in Farmland, and RVI in Meadow. Combination images generally outperform index images due to richer information across three channels. Though SentinelRGB is widely used for LULC, it does not surpass most combination images. Index images, being single-channel, contain less information, making them less effective than combination images.

\vspace{-0.5em}

\subsection{Bing RGB Image}

We describe the results from BingRGB, where the experiment includes training and testing using the full RGB image captured by the Bing satellite. The model with a full Bing RGB image outperforms the rest by significant margins. The average IoU is 0.44, and the average F1 score is 0.60. Not only in averages, but BingRGB also segments all five LULC classes more accurately compared to other models.

\vspace{-1.0em}

\section{Licensing and Ethical Considerations}

The dataset is available online at \url{https://doi.org/10.7910/DVN/LLR3RR} under the Creative Commons (CC) license. This license permits unrestricted use, distribution, and reproduction in any medium, provided that the original authors and source are credited.

\section{Conclusion}
\label{sec:conclusion}
This paper presents high-resolution pixel-wise Land Use Land Cover (LULC) ground truth annotations for the Dhaka metropolitan region, covering 4,392 square kilometers. The BOLM dataset includes eleven LULC classes across urban and rural areas. Baseline experiments using Sentinel 2A and Bing satellite imagery for five major LULC classes show that Bing’s high-resolution imagery (2.22m/pixel) consistently outperforms Sentinel 2A. However, Bing is limited to RGB channels and lacks free monthly availability in developing regions. Sentinel 2A, with multi-channel data including SWIR, remains valuable when high-resolution imagery is unavailable. Our findings highlight the trade-off between precision and accessibility, suggesting Bing/Google/ESA imagery for urban monitoring and Sentinel 2A for periodic applications like crop monitoring.

\vspace{-1.0em}

\section{Acknowledgements}

This research is partially supported by a grant from the Independent University, Bangladesh.

\bibliographystyle{IEEEbib}
\bibliography{strings,refs}

\newpage
\section*{Supplementary Materials}

\subsection{Analysis of Normalized Confusion Matrix}

\begin{figure*}
    \centering
    \begin{subfigure}[b]{0.50\textwidth}
        \centering
        \includegraphics[width=0.95\linewidth]{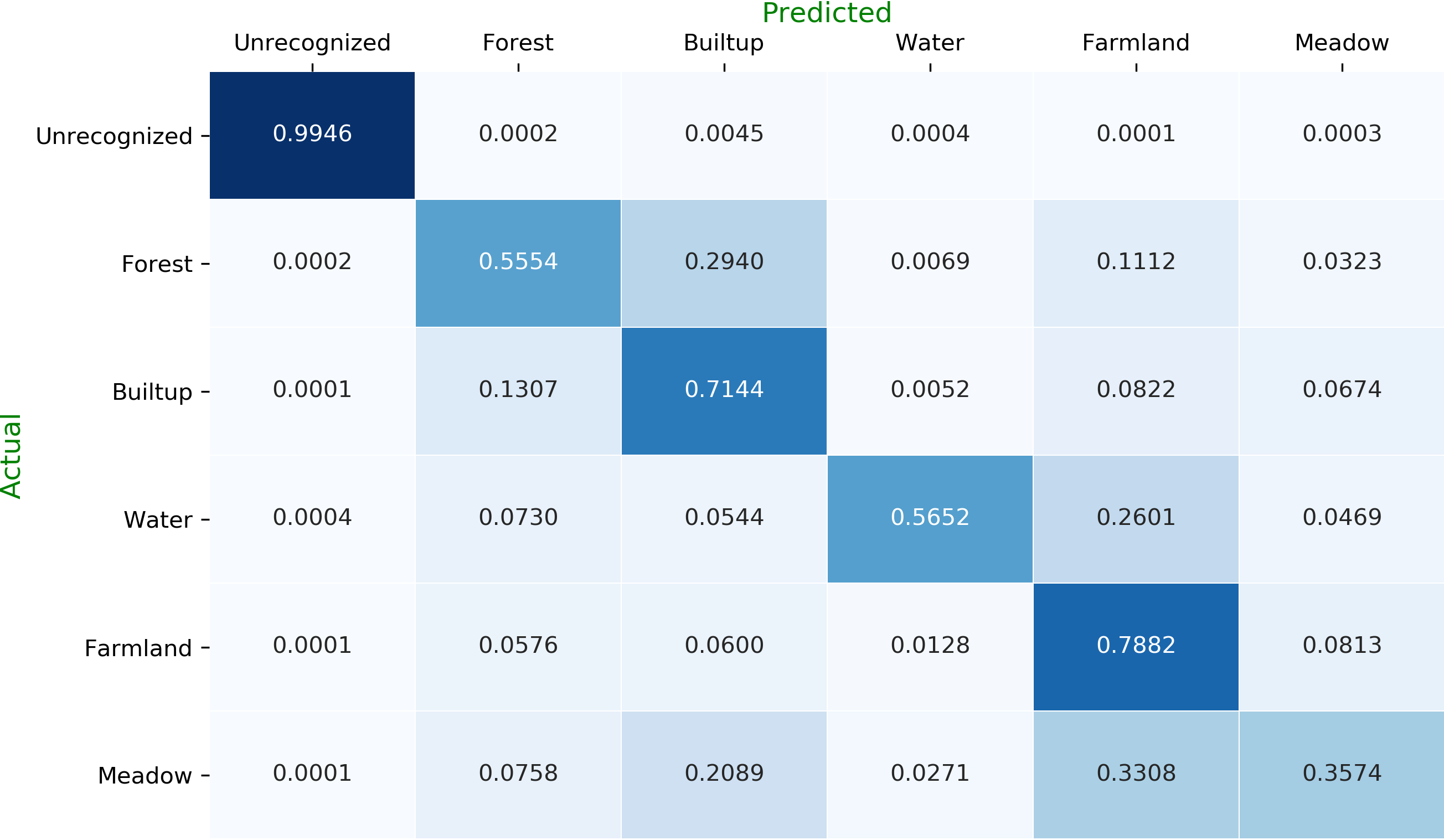}
        \caption{Bing RGB}
        \label{fig:conf_bing_RGB}
    \end{subfigure}%
    \begin{subfigure}[b]{0.50\textwidth}
        \centering
        \includegraphics[width=0.95\linewidth]{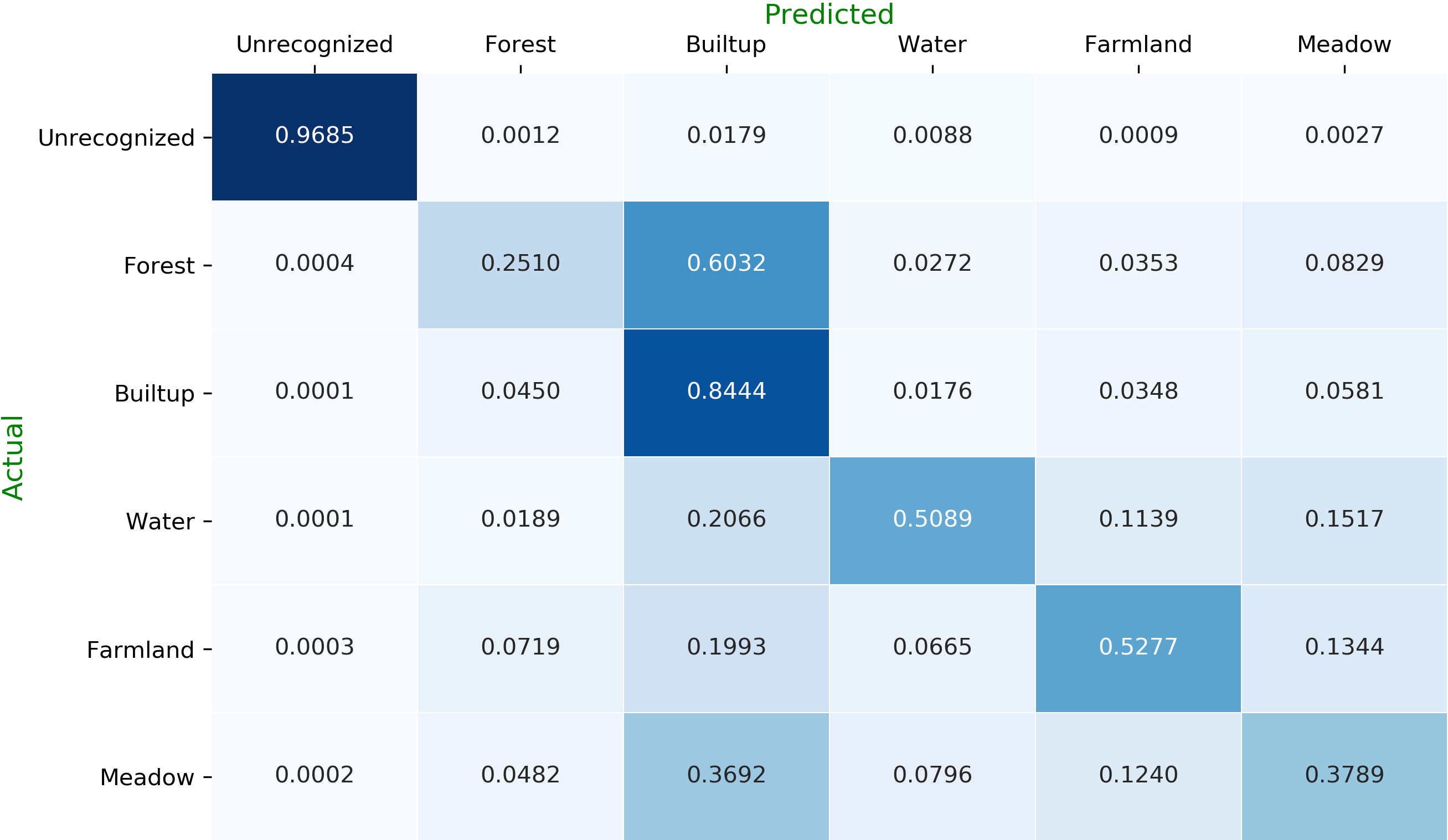}
        \caption{Sentinel-2 RGB}
        \label{fig:conf_senti_RGB}
    \end{subfigure}
    \caption{Normalized confusion matrix of Bing RGB and Sentinel-2 RGB images.}
    \label{fig:cm_bing_senti_rgb}
\end{figure*}

\begin{figure}[!h]
    \centering
    \includegraphics[width=\linewidth]{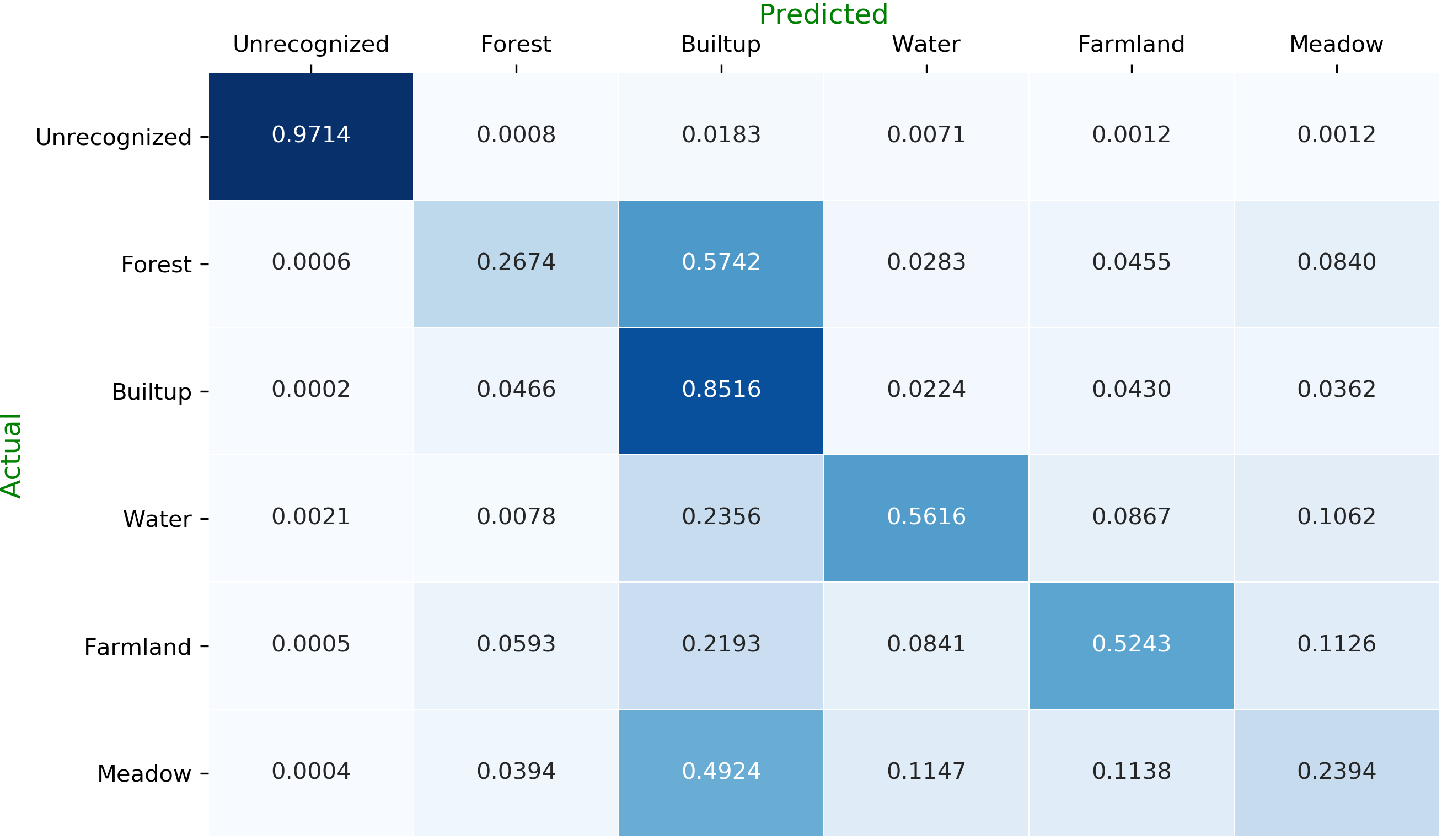}
    \caption{Normalized confusion matrix of Sentinel-2 FCI image}
    \label{fig:conf_sent_FCI}
\end{figure}

The normalized confusion matrices in Figures~\ref{fig:conf_bing_RGB}--\ref{fig:conf_sent_FCI} illustrate the performance of DeeplabV3+ on different datasets. BingRGB achieves the highest accuracy across all classes, with an average IoU of 0.44 and F1 of 0.60. The model struggles with inter-class confusions, particularly between Forest and Built-up areas, due to their proximity in urban settings. Water is often misclassified as Farmland due to similar colors in urban environments, and Meadow remains the most challenging class.

SentinelRGB achieves high accuracy for Built-up (84\%) but performs poorly in Forest (25\%) and Farmland (53\%) due to lower resolution. Sentinel-2 ATM improves Farmland and Water classification but struggles with Forest and Built-up, likely due to its spectral properties emphasizing atmospheric penetration. SWI achieves overall better performance than RGB but still struggles with Meadow misclassification due to resolution differences. FCI enhances Forest, Built-up, and Water segmentation but performs worst for Meadow, misclassifying it as Built-up due to spectral similarities in dry land areas.

Overall, combination images provide richer information, outperforming single-channel index images, while BingRGB delivers the highest segmentation accuracy.

\subsection{Annotation Validation and Data Reliability}  
To ensure the reliability of the annotation, we have followed a three-stage validation process.  

In stage 1, two groups of annotators independently annotated the dataset at the pixel level. A pixel-wise agreement matrix was calculated to compare their annotations (Table \ref{Stage-3pixeldisagree}). Disagreements were observed in all classes except the Brick and Road class. Notably, the Water class showed confusion with Farmland (3.4\%) and Marshland (5.66\%), as floating weeds in ponds and lakes often caused misclassification. The Farmland class also exhibited overlap with Roads, Rural Built-up areas, Marshland, and Meadow, particularly when crops were young or harvested.  
\begin{figure}[!htt]
    \centering
    \includegraphics[width=\linewidth]{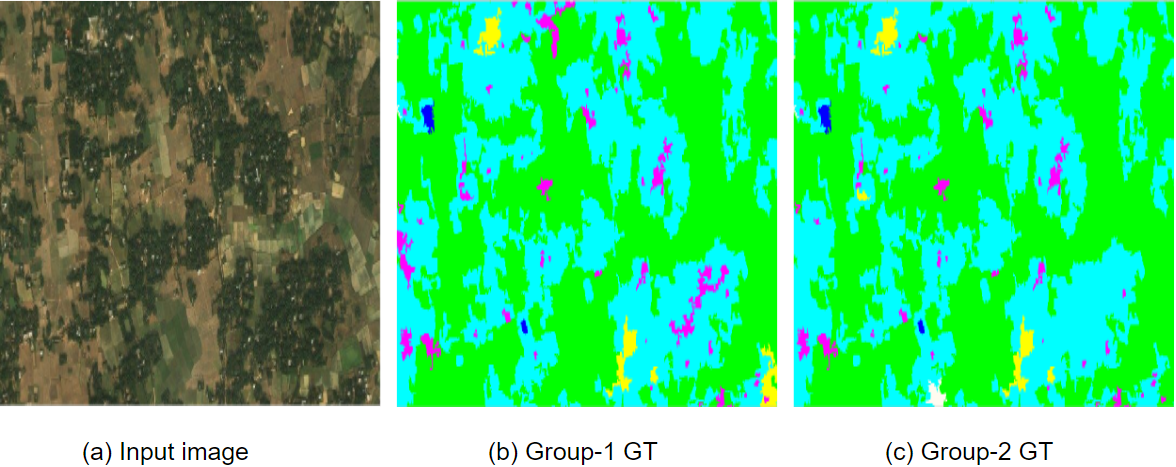}
    \caption{Disagreement in annotation between Group 1 and Group 2 members. Some of the water areas  have been missed by Group 2  }
    \label{disagree1}
\end{figure}

\begin{figure}[h]
    \centering
    \includegraphics[width=\linewidth]{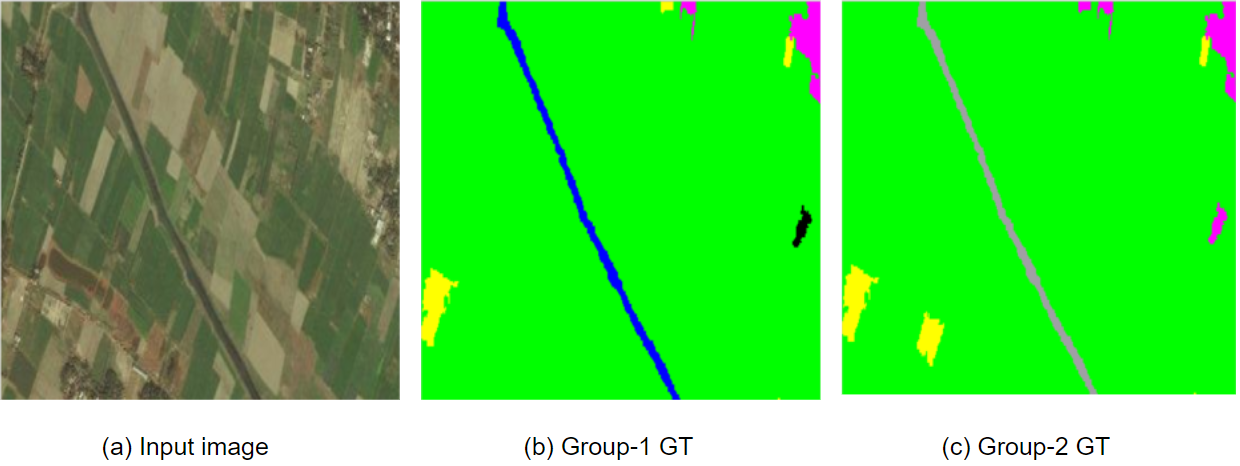}
    \caption{Disagreement in annotation between Group 1 and Group 2 members. One of the group 1 member annotated the long thin region as water (blue), whereas one group 2 member marked it as road. In Bangladesh, there are a lot of places where roads or rivers pass through vast farmland areas.}
    \label{disagree2}
\end{figure}

For the Forest class, minor confusion occurred mainly with Rural Built-up areas, as trees often surround houses in rural regions. This resulted in a 3.93\% disagreement. The highest agreement was observed for the Forest class (95.24\%), while the Road class had the lowest agreement (88.18\%). Figures \ref{disagree1} and \ref{disagree2} illustrate these disagreement issues. Many discrepancies were resolved in the second stage, with expert verification finalizing the annotations.

In stage 2, disagreement regions from stage 1 were isolated for detailed examination. Special attention was given to these regions (200 × 200 pixels), where annotator groups 1 and 2 disagreed. Each case was reviewed in discussions with the respective annotators. Using Photoshop on an iMac (24-inch screen), annotators manually refined the overlapping class boundaries at the pixel level.  

Tables 3 and 4 of the supplementary material show the revised agreement percentages. After stage 2, the disagreement percentage was significantly reduced, with the highest remaining disagreement at 4.2\%.

In stage 3, we sought validation from GIS experts with extensive experience in LULC annotations. Initially, the agreed-upon portions of the dataset—where both annotator groups reached a consensus—were selected. To further verify accuracy, 10\% of each class’s agreed portion was given to the expert group for evaluation. After expert annotation, comparisons showed that all classes except the Urban Structure class had above 99\% agreement, with Urban Structure at 98.585\%. Given this high level of agreement, the entire agreed portion from groups 1 and 2 was accepted without further verification. Table 5 of the supplementary material details the agreement percentages between the annotator groups and experts.  

Next, the areas of disagreement between group 1 and group 2 were fully annotated by the experts. Table 6 of the supplementary material presents the agreement/disagreement statistics between annotator disagreements and expert annotations. The expert-labeled pixels from Table 6 were adopted as final ground truth annotations. This stage marked the completion of the dataset, with the distribution of pixels across eleven classes illustrated in Figure \ref{fig:11classpercent}.

\begin{table*}[h!]
\caption{Stage-3: Group-1 and 2 disagreements vs. Expert annotation in number of pixels}
\label{Stage-3pixeldisagree}
\centering
\resizebox{\textwidth}{!}{%
\begin{tabular}{|r|l|l|r|r|r|r|r|r|r|r|r|r|r|}
\hline
\multicolumn{1}{|l}{} & & & \multicolumn{11}{c|}{Expert} \\ \cline{4-14}
\multicolumn{1}{|l|}{\begin{tabular}[c]{@{}l@{}}Total Pixels Selected \\ for Validation\\ (100\% of 1 and 2 \\ Disagree)\end{tabular}} & & &
\multicolumn{1}{l|}{Unrecognized} & \multicolumn{1}{l|}{Water} & \multicolumn{1}{l|}{Farmland} & \multicolumn{1}{l|}{Forest} & \multicolumn{1}{l|}{\begin{tabular}[c]{@{}l@{}}Urban\\Structure\end{tabular}} & \multicolumn{1}{l|}{\begin{tabular}[c]{@{}l@{}}Brick\\Factory\end{tabular}} & \multicolumn{1}{l|}{Road} & \multicolumn{1}{l|}{\begin{tabular}[c]{@{}l@{}}Urban\\Builtup\end{tabular}} & \multicolumn{1}{l|}{\begin{tabular}[c]{@{}l@{}}Rural\\Builtup\end{tabular}} & \multicolumn{1}{l|}{Marshland} & \multicolumn{1}{l|}{Meadow} \\ \hline
\multicolumn{1}{|r|}{244,452} &
\multirow{11}{*}{\rotatebox[origin=c]{90}{Both Group 1 and Group 2 Disagree}} &
Unrecognized &
\multicolumn{1}{r|}{243,437} & \multicolumn{1}{r|}{37} & \multicolumn{1}{r|}{49} & \multicolumn{1}{r|}{244} & \multicolumn{1}{r|}{24} & \multicolumn{1}{r|}{0} & \multicolumn{1}{r|}{0} & \multicolumn{1}{r|}{122} & \multicolumn{1}{r|}{244} & \multicolumn{1}{r|}{49} & \multicolumn{1}{r|}{244} \\ \cline{1-1} \cline{3-14}
\multicolumn{1}{|r|}{1,058,694} & & Water &
\multicolumn{1}{r|}{106} & \multicolumn{1}{r|}{1,049,695} & \multicolumn{1}{r|}{0} & \multicolumn{1}{r|}{0} & \multicolumn{1}{r|}{212} & \multicolumn{1}{r|}{529} & \multicolumn{1}{r|}{0} & \multicolumn{1}{r|}{1,588} & \multicolumn{1}{r|}{5,293} & \multicolumn{1}{r|}{1,059} & \multicolumn{1}{r|}{212} \\ \cline{1-1} \cline{3-14}
\multicolumn{1}{|r|}{9,250,035} & & Farmland &
\multicolumn{1}{r|}{925} & \multicolumn{1}{r|}{1,850} & \multicolumn{1}{r|}{9,172,335} & \multicolumn{1}{r|}{32,375} & \multicolumn{1}{r|}{9,250} & \multicolumn{1}{r|}{13,875} & \multicolumn{1}{r|}{0} & \multicolumn{1}{r|}{0} & \multicolumn{1}{r|}{3,700} & \multicolumn{1}{r|}{11,100} & \multicolumn{1}{r|}{4,625} \\ \cline{1-1} \cline{3-14}
\multicolumn{1}{|r|}{1,901,067} & & Forest &
\multicolumn{1}{r|}{380} & \multicolumn{1}{r|}{190} & \multicolumn{1}{r|}{6,844} & \multicolumn{1}{r|}{1,878,254} & \multicolumn{1}{r|}{3,802} & \multicolumn{1}{r|}{951} & \multicolumn{1}{r|}{2,852} & \multicolumn{1}{r|}{0} & \multicolumn{1}{r|}{4,753} & \multicolumn{1}{r|}{2,471} & \multicolumn{1}{r|}{570} \\ \cline{1-1} \cline{3-14}
\multicolumn{1}{|r|}{169,327} & & \begin{tabular}[c]{@{}l@{}}Urban\\Structure\end{tabular} &
\multicolumn{1}{r|}{254} & \multicolumn{1}{r|}{0} & \multicolumn{1}{r|}{847} & \multicolumn{1}{r|}{169} & \multicolumn{1}{r|}{165,906} & \multicolumn{1}{r|}{85} & \multicolumn{1}{r|}{339} & \multicolumn{1}{r|}{508} & \multicolumn{1}{r|}{339} & \multicolumn{1}{r|}{847} & \multicolumn{1}{r|}{34} \\ \cline{1-1} \cline{3-14}
\multicolumn{1}{|r|}{47,042} & & \begin{tabular}[c]{@{}l@{}}Brick\\Factory\end{tabular} &
\multicolumn{1}{r|}{0} & \multicolumn{1}{r|}{235} & \multicolumn{1}{r|}{0} & \multicolumn{1}{r|}{0} & \multicolumn{1}{r|}{24} & \multicolumn{1}{r|}{46,680} & \multicolumn{1}{r|}{0} & \multicolumn{1}{r|}{56} & \multicolumn{1}{r|}{47} & \multicolumn{1}{r|}{0} & \multicolumn{1}{r|}{0} \\ \cline{1-1} \cline{3-14}
\multicolumn{1}{|r|}{185,121} & & Road &
\multicolumn{1}{r|}{19} & \multicolumn{1}{r|}{0} & \multicolumn{1}{r|}{0} & \multicolumn{1}{r|}{0} & \multicolumn{1}{r|}{185} & \multicolumn{1}{r|}{0} & \multicolumn{1}{r|}{183,992} & \multicolumn{1}{r|}{926} & \multicolumn{1}{r|}{0} & \multicolumn{1}{r|}{0} & \multicolumn{1}{r|}{0} \\ \cline{1-1} \cline{3-14}
\multicolumn{1}{|r|}{1,532,498} & & \begin{tabular}[c]{@{}l@{}}Urban\\Builtup\end{tabular} &
\multicolumn{1}{r|}{613} & \multicolumn{1}{r|}{0} & \multicolumn{1}{r|}{0} & \multicolumn{1}{r|}{0} & \multicolumn{1}{r|}{6,130} & \multicolumn{1}{r|}{4,597} & \multicolumn{1}{r|}{10,727} & \multicolumn{1}{r|}{1,510,276} & \multicolumn{1}{r|}{153} & \multicolumn{1}{r|}{0} & \multicolumn{1}{r|}{0} \\ \cline{1-1} \cline{3-14}
\multicolumn{1}{|r|}{2,495,021} & & \begin{tabular}[c]{@{}l@{}}Rural\\Builtup\end{tabular} &
\multicolumn{1}{r|}{3,493} & \multicolumn{1}{r|}{2,246} & \multicolumn{1}{r|}{3,743} & \multicolumn{1}{r|}{1,248} & \multicolumn{1}{r|}{0} & \multicolumn{1}{r|}{0} & \multicolumn{1}{r|}{0} & \multicolumn{1}{r|}{749} & \multicolumn{1}{r|}{2,483,544} & \multicolumn{1}{r|}{0} & \multicolumn{1}{r|}{0} \\ \cline{1-1} \cline{3-14}
\multicolumn{1}{|r|}{322,169} & & Marshland &
\multicolumn{1}{r|}{32} & \multicolumn{1}{r|}{1,289} & \multicolumn{1}{r|}{967} & \multicolumn{1}{r|}{644} & \multicolumn{1}{r|}{0} & \multicolumn{1}{r|}{0} & \multicolumn{1}{r|}{0} & \multicolumn{1}{r|}{0} & \multicolumn{1}{r|}{0} & \multicolumn{1}{r|}{319,108} & \multicolumn{1}{r|}{129} \\ \cline{1-1} \cline{3-14}
\multicolumn{1}{|r|}{3,149,345} & & Meadow &
\multicolumn{1}{r|}{1,575} & \multicolumn{1}{r|}{1,249} & \multicolumn{1}{r|}{7,495} & \multicolumn{1}{r|}{3,123} & \multicolumn{1}{r|}{0} & \multicolumn{1}{r|}{0} & \multicolumn{1}{r|}{0} & \multicolumn{1}{r|}{0} & \multicolumn{1}{r|}{3,123} & \multicolumn{1}{r|}{11,868} & \multicolumn{1}{r|}{6,216,219} \\ \hline
\end{tabular}%
}
\end{table*}

\begin{figure*}[htbp]
    \centering
    \includegraphics[width=.75\textwidth]{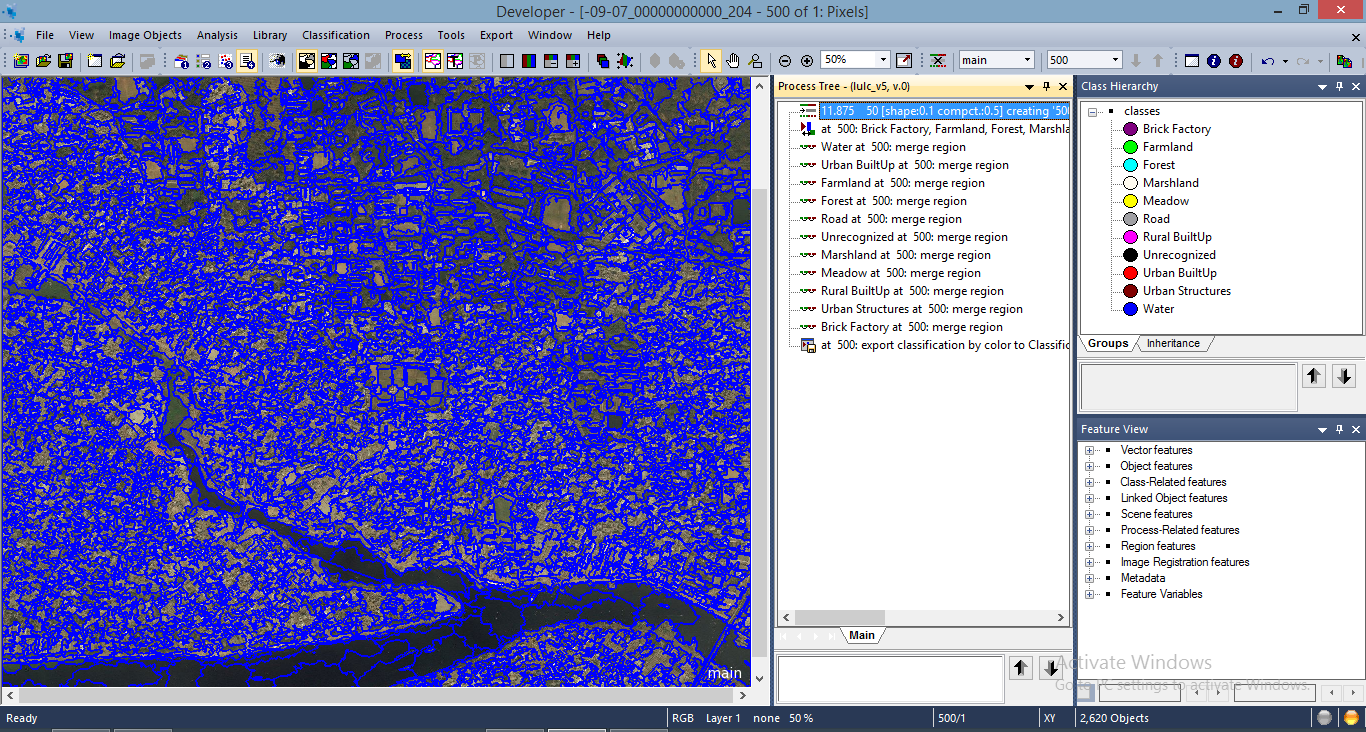}
    \caption{Segmenting image into small polygons}
    \label{fig:2}
\end{figure*}

\begin{figure*}[htbp]
    \centering
    \includegraphics[width=.750\textwidth]{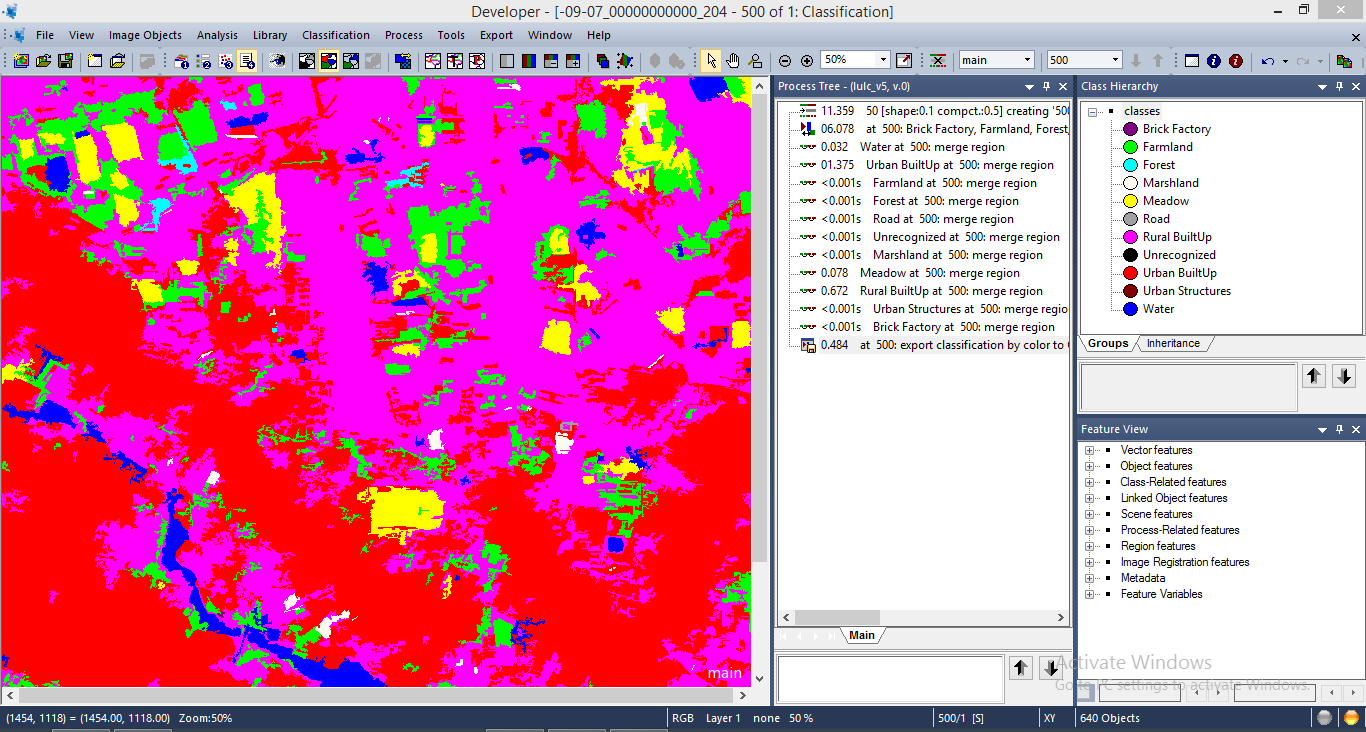}
    \caption{Exporting after fully labeling the ground truth}
    \label{fig:4}
\end{figure*}

\begin{figure*}
    \centering
    \begin{subfigure}[b]{.50\textwidth}
        \centering
        \includegraphics[width=\linewidth]{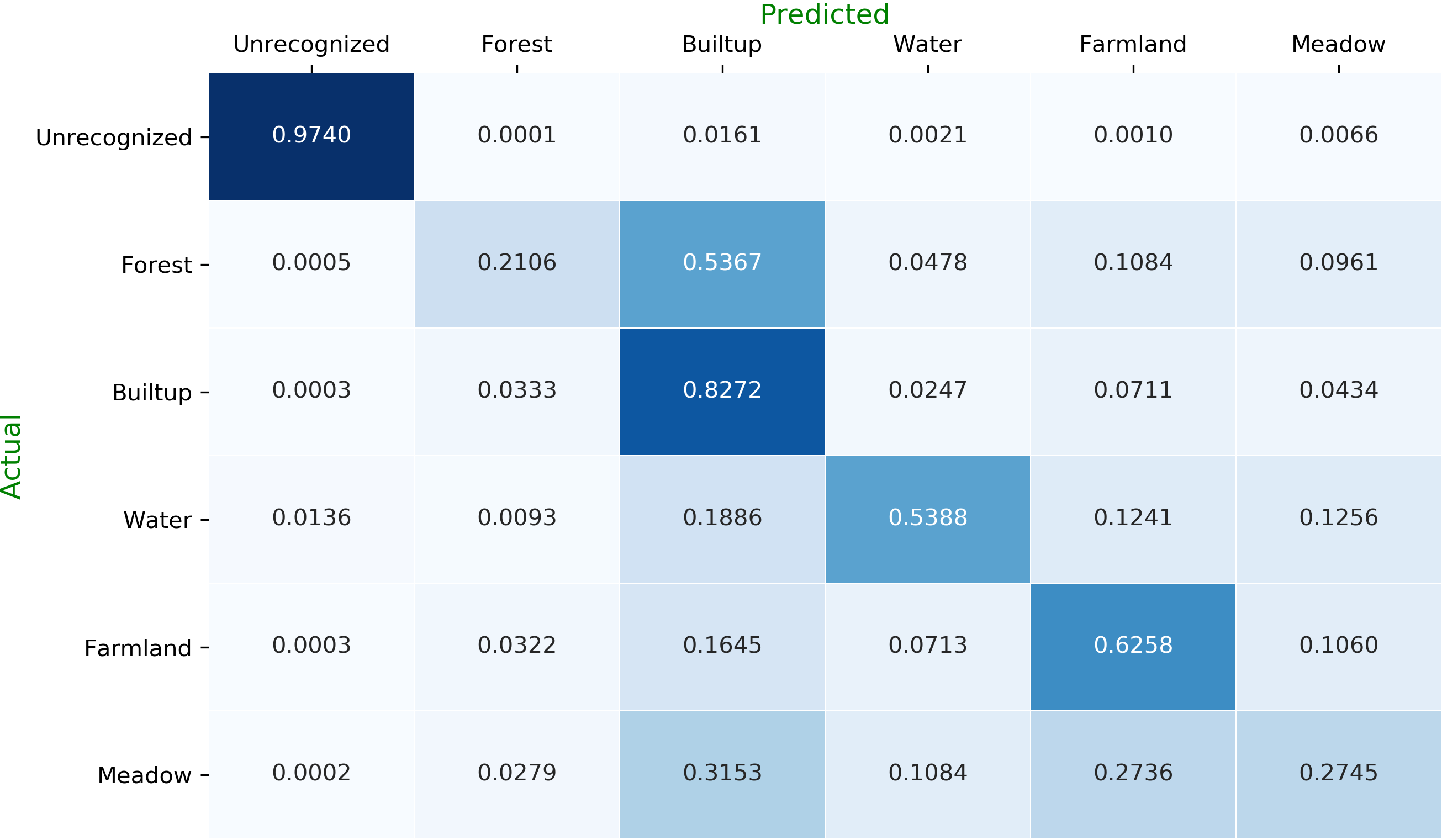}
        \caption{Sentinel-2 ATM}
        \label{fig:conf_sent_atm}
    \end{subfigure}%
    \begin{subfigure}[b]{.50\textwidth}
        \centering
        \includegraphics[width=\linewidth]{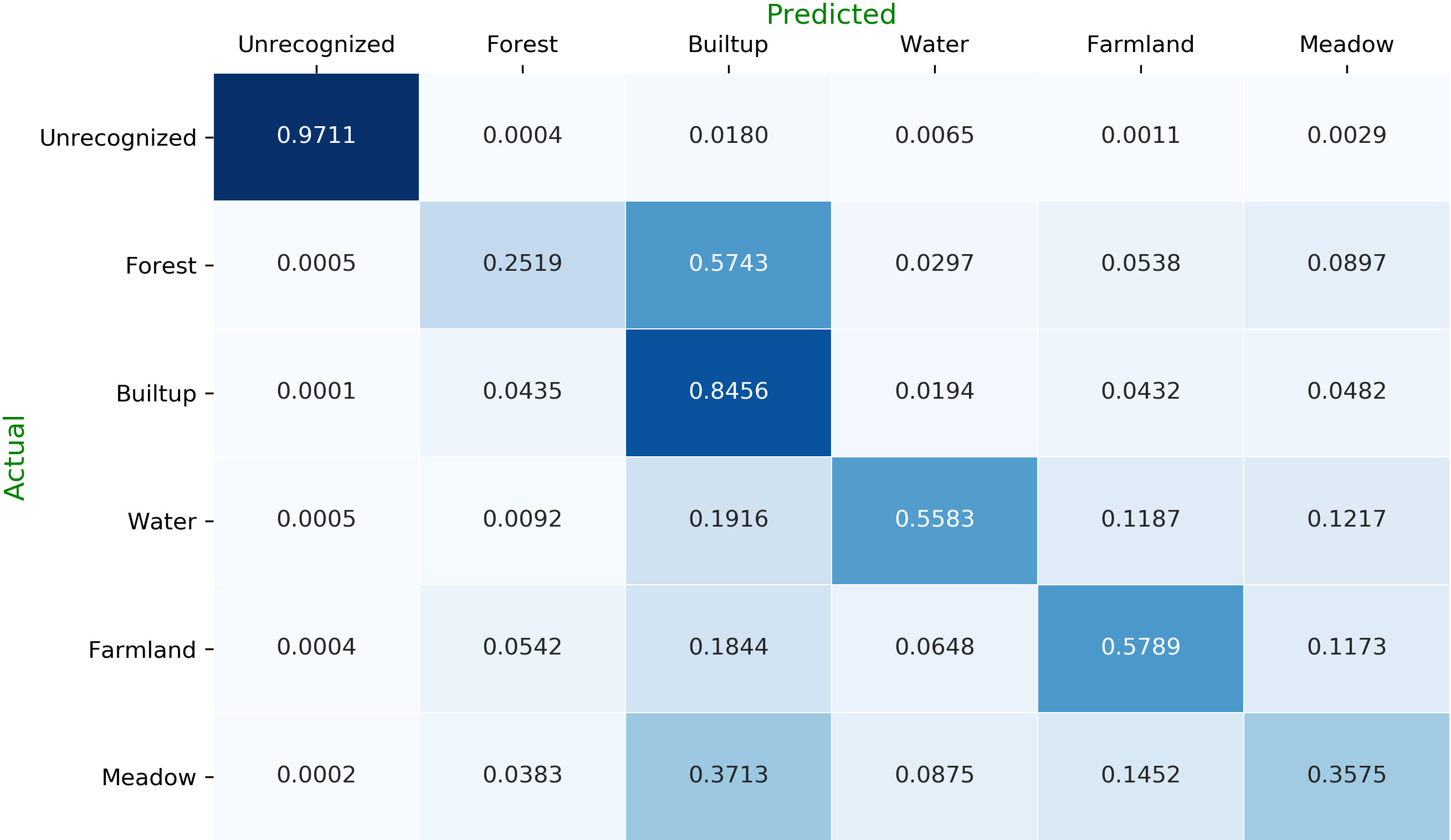}
        \caption{Sentinel-2 SWIR}
        \label{fig:conf_sent_swir}
    \end{subfigure}
    \caption{Normalized confusion matrix of Sentinel-2 ATM and SWIR combination images.}
    \label{fig:cm_atm_swir}
\end{figure*}

\end{document}